\documentclass{article}

\usepackage[preprint]{nips_2018}

\usepackage{natbib}
\usepackage[utf8]{inputenc} 
\usepackage[T1]{fontenc}    
\usepackage{hyperref}       
\usepackage{url}            
\usepackage{booktabs}       
\usepackage{amsfonts}       
\usepackage{nicefrac}       
\usepackage{microtype}      

\usepackage{graphicx} 
\usepackage{algorithm} 
\usepackage{algorithmicx} 
\usepackage{algpseudocode}
\usepackage{amsmath}
\usepackage{subfigure}

\title{Object-Level Representation Learning for Few-Shot Image Classification}

\author{
  Liangqu Long$^\dagger$, Wei Wang$^\dagger$, Jun Wen$^\ddagger$, Meihui  Zhang$^\diamond$, Qian  Lin$^\dagger$,  Beng Chin  Ooi$^\dagger$\\
  $^\dagger$National University of Singapore, $^\ddagger$Zhejiang University, $\diamond$Beijing Institute of Technology\\ 
  \texttt{$^\dagger$dcslong@nus.edu.sg, $^\dagger$\{wangwei, linqian, ooibc\}@comp.nus.edu.sg, }\\
  \texttt{$^\ddagger$jungel2star@gmail.com, $\diamond$meihui\_zhang@bit.edu.cn} \\
}

\begin{document}

\maketitle

\begin{abstract}
Few-shot learning that trains image classifiers over few labeled examples per category is a challenging task. In this paper, we propose to exploit an additional big dataset with different categories to improve the accuracy of few-shot learning over our target dataset. Our approach is based on the observation that images can be decomposed into objects, which may appear in images from both the additional dataset and our target dataset. We use the object-level relation learned from the additional dataset to infer the similarity of images in our target dataset with unseen categories. Nearest neighbor search is applied to do image classification, which is a non-parametric model and thus does not need fine-tuning. We evaluate our algorithm on two popular datasets, namely Omniglot and MiniImagenet. We obtain 8.5\% and 2.7\% absolute improvements for 5-way 1-shot and 5-way 5-shot experiments on MiniImagenet, respectively. Source code will be published upon acceptance.

\end{abstract}

\section{Introduction}
Real-world data typically follows power-law distributions, where the majority of the data categories have only a small number of examples. 
For instance, to train an image classifier for food images, one would probably crawl few images for some local dishes and many images for common dishes like burger. 
Similarly, there are few images for new products, e.g. new toys. 
However, state-of-the-art image classifiers, i.e. deep convolutional neural networks (ConvNets)~\citet{DBLP:conf/nips/KrizhevskySH12}, are extremely hungry for data. 
The popular benchmark datasets for ConvNets, including CIFAR10 and ImageNet~\cite{imagenet_cvpr09}, usually have more than $1000$ images per category. 
Fine-tuning ConvNets ~\cite{DBLP:journals/corr/YosinskiCBL14} by transferring the knowledge (i.e. parameters) learned from a big dataset could alleviate the gap in some degree, but still fails to resolve the issue. 
One reason is that the widely used gradient-based optimization algorithms need many iterations over plenty of examples to adapt the ConvNets (with a large number of parameters) for new categories \cite{ravi2016optimization}. 

Two approaches have been proposed towards the above challenge. 
They are referred as \textit{few-shot learning}, which trains classifiers over datasets with few (e.g. less than $20$) examples per category. 
The first approach~\cite{ravi2016optimization,DBLP:journals/corr/LiZCL17,DBLP:journals/corr/FinnAL17} is based on meta-learning. 
MAML~\cite{DBLP:journals/corr/FinnAL17} trains a meta-learner to provide good initialization for the parameters of the classifier. Meta-SGD~\cite{DBLP:journals/corr/LiZCL17}'s meta-learner generates adaptive learning rate for training the classifier. \cite{ravi2016optimization} replaces the gradient-based optimizer with a LSTM to train the classifier.
The second approach~\cite{kochsiamese,DBLP:journals/corr/VinyalsBLKW16,DBLP:journals/corr/SantoroBBWL16,DBLP:journals/corr/SnellSZ17,DBLP:journals/corr/abs-1711-06025} is based on metric learning. 
It learns an embedding function to project the images into an embedding space and then classify images from new categories via nearest neighbour search (NNS). 
No fine-tuning is required over the new categories as the NNS classifier is non-parametric. 
The embedding functions are vital to the classification accuracy, which must be general enough to extract effective features for measuring the distance or similarity between images from the unseen categories. 

\begin{figure}[htb]
    \centering
    \includegraphics[width=0.4\textwidth]{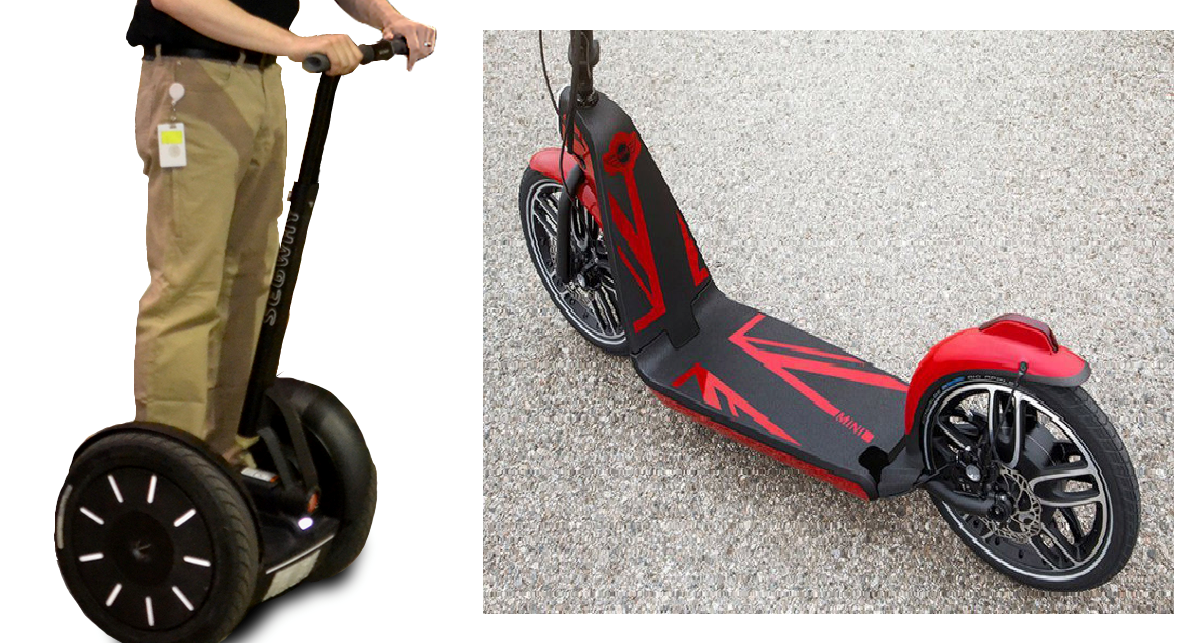}
    \caption{Left: Segway, Right: electric scooter. Human can easily recognize Segway as one type of motors by its components: wheels, pedals, rider.}
    \label{fig:segway}
\end{figure}


In this paper, we follow the second approach and take the following observation into consideration. 
Humans usually recognize a new (unseen) object by decomposing it and comparing the components with those of other objects they have seen before. Take the Segway in Figure~\ref{fig:segway} as an example (\citet{lake15science}), 
although Segway could be new to us, we are familiar with its components (i.e. objects), e.g. wheels and pedals, which are similar to those of the motors or electric scooters in our memory. 
By comparing the components of Segway and electric scooters, we would know Segway is a kind of motors for riding. 

Motivated by human intelligence, we propose a novel approach by learning image similarity based on their object-level relation. This approach is called OLFSL, short for Object-Level Few-Shot Learning. OLFSL compares the objects from two images to learn the object-level relations, which are used to infer the image-level similarity. More specifically, OLFSL is composed of three modules: representation learning $\mathcal{F}_{\Phi}(x)$, objects relation learning $\mathcal{R}_{\theta}(a, b)$ and similarity learning $\mathcal{S}_{\phi}(r)$. $\mathcal{F}_{\Phi}$ extracts features of objects from each image; The object-level relations of two images are learned by feeding the object features into $\mathcal{R}_{\theta}(a, b)$. All learned relations from $\mathcal{R}_{\theta}(a, b)$ are aggregated to generate the image-level similarity score: $\mathcal{S}_{\phi}(\cdot) \mapsto s_{ab}$. The three modules are trained using the additional big dataset. Nearest neighbor search is applied over the target dataset to do few-shot classification.

The primary contribution of this work is exploiting the object-level relation learned from known categories (the additional dataset) to infer the similarity of samples from unseen categories (the target dataset) for few-shot learning.
We evaluate our approach on two popular datasets, Omniglot~\citet{lake15science}, MiniImagenet~\citet{DBLP:journals/corr/VinyalsBLKW16}. The experimental results show that we reach the state-of-the-art performance on Omniglot. For MiniImagenet, we achieve 8.5\% and 2.7\% absolute improvements over the state-of-the-art methods for 5-way 1-shot and 5-way 5-shot experiments respectively. Besides,  OLFSL is model-agnostic, scalable, and free of fine-tuning on new tasks.

\begin{figure*}[htb]
	\centering
	\includegraphics[width=0.95\textwidth]{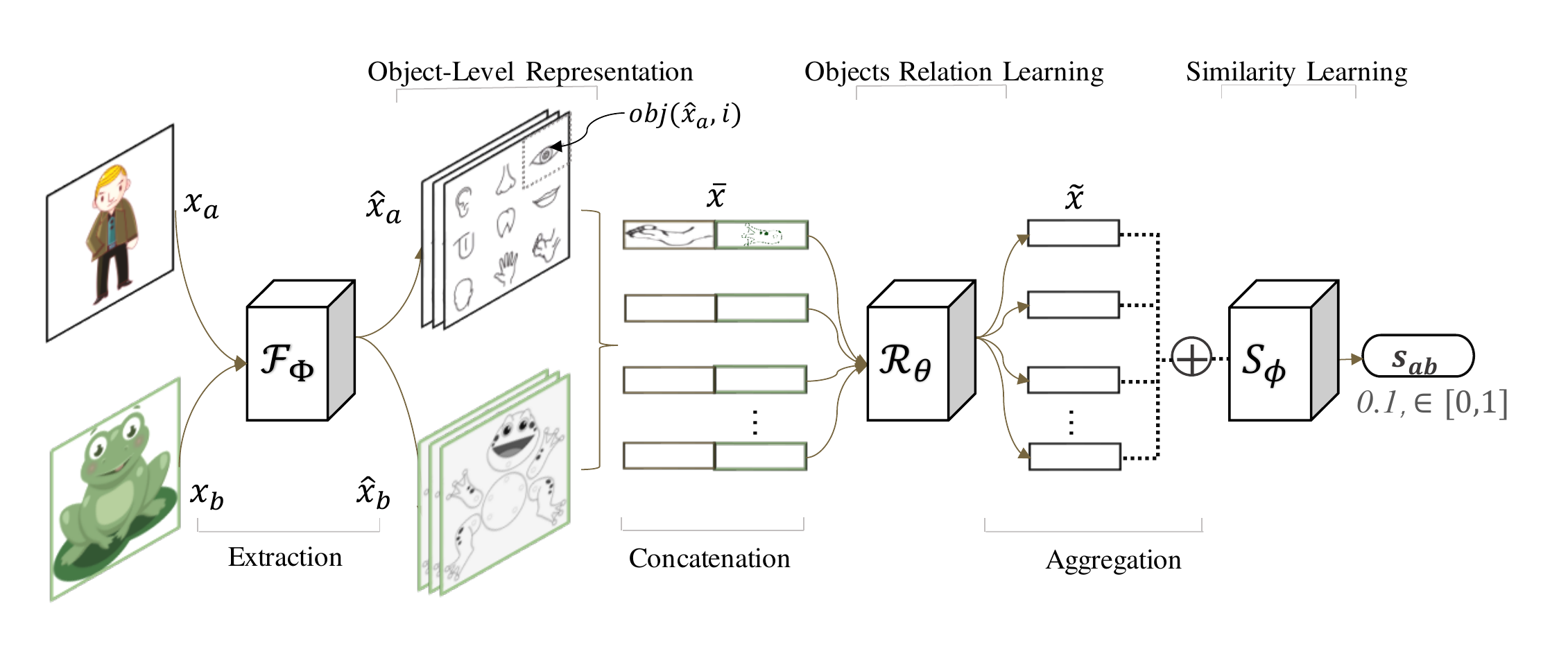}
	\caption{OLFSL architecture. Image $a$ from the support set and $b$ from the query set are fed into $\mathcal{F}_{\Phi}(x)$ separately. $\{obj(\hat{x}_a, i)\}_{i=1}^{d*d}$ (resp. $\{obj(\hat{x}_b, j)\}_{j=1}^{d*d}$) denotes the object representations extracted from the raw image feature $x_a$ (resp. $x_b$). Objects from $\hat{x}_a$ are concatenated with objects in $\hat{x}_b$ pair-wisely. The concatenated vectors are fed into $\mathcal{R}_{\theta}$ network to learn object-level relations. All object-level relations are aggregated together to learn the similarity score by $\mathcal{S}_{\phi}\mapsto s_{ab}$, $s_{ab}\in [0, 1]$. 
	\label{fig:model}}
\end{figure*}

\section{Methodology}

In this section, we first give the problem definition of few-shot learning, and then introduce the model architecture, and finally explain the details about the training and inference procedure. We use 1-shot learning to explain the idea and then extend it to K-shot learning (K>1).


\subsection{Problem Definition}

For few-shot image learning, we are given a support set of labeled images $\mathcal{E}_{spt}=\{(x_i, y_i)\}_{i=1}^{N}$, where $x_i\in R^{D}$ is the feature of an image, and $y_i\in \{0, 1, 2, \cdots, n-1\}$ is the label. 
If the number of images per category is $K$ and the classes number is $N$, the task is denoted as N-way K-shot learning, which classifies the images from a query set (denoted as $\mathcal{E}_{qry}$) by assigning each image with a label from $\{0, 1, 2, \cdots, N-1\}$. 
The support set $\mathcal{E}_{spt}$ and query set $\mathcal{E}_{qry}$ together form a training (or test) episode 
Typically, $K$ is a small number for few-shot setting, e.g. $K=1$ or $5$.

\subsection{Model Architecture}

Our model consists of three parts as shown in Figure~\ref{fig:model}. The architecture configuration of each part is illustrated in Figure~\ref{fig:archo}.

\subsubsection{Object Feature Learning}
For each image sample $a$, we denote $x_a$ as the raw image feature  and $y_a$ as the label. $x_a$ is fed into a convolutional neural network $\mathcal{F}_{\Phi}$ to extract the object features. In particular, we decompose the feature maps $\hat{x}\in R^{d*d*c}$  from the last convolution layer to get the object features. There are in total $d * d$ objects, i.e. $\{obj(\hat{x}, i)\in R^{1*1*c}\}_{i=1}^{d*d}$, where $d$ stands for the number of objects on horizontal and vertical dimension (we assume $a$ has equal height and width). The i-th object is denoted as $obj(\hat{x}, i)\in R^{1*1*c}, i\in \{0, 1, 2,... d*d-1\}$.

\subsubsection{Object-Level Relation Learning}
This module learns the object-level relations from object-level representations extracted from $\mathcal{F}_{\Phi}$. Give two images $a\in \mathcal{E}_{spt}$ and $b\in \mathcal{E}_{qry}$, $\mathcal{F}_{\Phi}$ extracts the object-level representations as $\hat{x}_a$ and $ \hat{x}_b$ respectively. 
We compare $\hat{x}_a$ and $ \hat{x}_b$ following rule $\mathcal{C}$. A simple rule is to concatenate objects from $\hat{x}_a$ and $\hat{x}_b$ pair-wisely, i.e. 

\begin{equation} 
    \label{formu:Comb2}
    \mathcal{C}(\hat{x}_a, \hat{x}_b)=\{concat(obj(\hat{x}_a, i), obj(\hat{x}_b, j))\}_{i=1, j=1}^{i=d*d, j=d*d}
\end{equation}

Other rules can also be applied, e.g. concatenating objects at the same spatial location.


Each concatenated vector is fed into another fully connected neural network to learn the object-level relation. The output feature vector, denotd as $\tilde{x}_k\in R^{\tilde{c}}$,  represents the relation between $obj(\hat{x}_a, i)$ and $obj(\hat{x}_b, j)$. In total, there are $d^4$ such relation feature vectors.


All object relation vectors $\{\tilde{x}_i\}_{i=1}^{d^4}$ are aggregated so as to get the image-level relation,
\begin{equation}
    \label{formu:merge}
    m_{ab} =  \tilde{x}_0 \oplus  \tilde{x}_1 \oplus  \tilde{x}_2 \oplus ... \oplus  \tilde{x}_{d^4-1} \in R^{\tilde{c}}
\end{equation}
where $\oplus$ stands for element-wise add operation.


\subsubsection{Similarity Learning}

The image-level relation feature vector $m_{ab}$ is fed into a fully connected neural network to generate the final image-level similarity score,

\begin{equation}
    \label{formu:score}
    s_{ab} = \mathcal{S}_{\phi}(m_{ab})
\end{equation} 
where $s_{ab}$ is the similarity between two samples $a$ and $b$. It is normalized to $[0, 1]$, where 0 stands for completely distinct and 1 for almost the same.

\subsection{Training and Inference}


We design the training procedure following \cite{DBLP:journals/corr/VinyalsBLKW16} to make the training and inference conditions match. More specifically, We divide the whole dataset into there parts, namely $\mathcal{D}_{train}$, $\mathcal{D}_{val}$ and $\mathcal{D}_{test}$ with disjoint label space. $\mathcal{D}_{train}$ serves as the additional big dataset to learn the object-level relation. For each sub dataset, we create a N-way K-shot episode (or task), denoted as $\Gamma_i$, by randomly sampling images to construct the support set and query set. Cross-entropy is employed as the training objective as shown in Equation~\ref{formu:loss}, where the ground truth y=1 if a and b are from the same category; otherwise the ground truth is y=0. By optimizing this objective, the predicted similarity becomes close to the ground truth similarity.
\begin{equation}
    \label{formu:loss}
    \Phi, \theta, \phi = \underset{\Phi, \theta, \phi}{argmax}\sum_{x_a, x_b \sim \Gamma_i}{y*\log P(1|x_a, x_b) + (1-y)*(1-\log P(0|x_a, x_b))}
\end{equation} 

During inference, for each query, we compare it with each image from the support set by feeding them through the model. Nearest neighbor search then assigns the label of the image with the largest similarity to the query. Fine-tuning is not required for OLFSL.


For the case of $K>1$, there are more than one example per category in the support set. Feature maps for images from the same category are averaged to get an average representation to reduce the representation variance,

\begin{equation}
    \label{formu:ensemble}
    \hat{x}^{avg} = \sum_{i=0}^{K-1}(\hat{x}_i) / K
\end{equation}

The averaged representation is also used in Prototypical Network~\cite{DBLP:journals/corr/SnellSZ17}. 
Other modules for K-shot learning work in the same way as one-shot learning. 
The pseudo code of our algorithm is shown in Algorithm~\ref{alg:OLFSL}.

\begin{algorithm}
	\caption{OLFSL Pseudo-code for training and testing\label{alg:OLFSL}}
	\begin{algorithmic}[1]	 
	    \State \Comment{Training Stage}
	    \State randomly initialize $\Phi, \theta, \phi$
	    \While{not done}
                \State sample one episode
	            \State randomly sample $\{(x_a, y_a), (x_b, y_b)\}$ from $\mathcal{E}_{spt}$ and $\mathcal{E}_{qry}$
	            \State extract object-level representation $\hat{x}_a, \hat{x}_b$
	            \State compute similarity score $s_{ab}$ via Equation~\ref{formu:score}
	            \State compute Equation~\ref{formu:loss} and update parameters $\Phi, \theta, \phi$ via SGD
	    \EndWhile
	    
	    \State \Comment{Testing Stage}
	    \State Given the support set $\mathcal{E}_{spt}$ and one query sample $b \in R^{D}$
	    \For{each sample $a$ in $\mathcal{E}_{spt}$}
	        \State compute similarity score $s_{ab}$ via Equation~\ref{formu:score}
	        \State $s_{ab} \rightarrow \mathcal{S}$
	    \EndFor
	    \State Return the label with the largest (aggregated) similarity score.
	\end{algorithmic}
\end{algorithm}

\section{Experimental Evaluation}
We evaluate our approach on two popular benchmark datasets: Omniglot, MiniImagenet. For each dataset, we partition it into training, validation and testing subsets: $\mathcal{D}_{val}$, $\mathcal{D}_{train}$ and $\mathcal{D}_{test}$. 
The N-way K-shot learner is trained by sampling $N$ classes and $K$ examples per class for each training episode $\mathcal{E}$. We introduce the experiments on the two datasets respectively in the following subsections.

\subsection{Evaluation on Omniglot}

Omniglot~\cite{Lake1332} contains 1623 characters (classes) from 50 different alphabets. Each class has 20 samples drawn by different people. We use 1200 classes for training (including validation), and the remaining 423 classes for testing. Following \cite{DBLP:journals/corr/abs-1711-06025,DBLP:journals/corr/SnellSZ17}, all input images are augmented by rotations in multiples of 90 degrees. In every testing episode, 15 query images per class are tested.

\begin{table*}[htb!]
\centering
\caption{Performance comparison on Omniglot dataset. \textit{FT} means fine-tuned or not.}
\label{tb:omniglot}
\begin{tabular}{@{} lccccc @{}}

\toprule
\hline

\textbf{MODEL} &
\textbf{FT} & 
\multicolumn{2}{c}{\textbf{5-WAY }} &
\multicolumn{2}{c}{\textbf{20-WAY }} \\

& \textbf{} & 1-shot & 5-shot & 1-shot &5-shot  \\
\hline
\midrule 

\textsc{\cite{kochsiamese}} &N &96.7\% &98.4\% &88.0\% &96.5\% \\ 
\textsc{\cite{kochsiamese}}  &Y &97.3\% &98.4\% &88.1\% &97.0\% \\ 
\textsc{MANN}~\cite{DBLP:journals/corr/SantoroBBWL16} &N & 82.8\% & 94.9\% & - &- \\
\textsc{\cite{DBLP:journals/corr/VinyalsBLKW16}}  &N & 98.1\% & 98.9\% &93.8\% & 98.5\% \\
\textsc{\cite{DBLP:journals/corr/VinyalsBLKW16}}  &Y &97.9\% &98.7\% &93.5\% &98.7\% \\ 
\textsc{\cite{DBLP:journals/corr/KaiserNRB17}}  &N &98.4\% &99.6\% &95.0\% &98.6\% \\
\textsc{\cite{DBLP:journals/corr/MunkhdalaiY17}}  &N &99.0\% & - & 97.0\% & -\\ 
\textsc{\cite{DBLP:journals/corr/SnellSZ17}}  &N &98.8\% &99.7\%  &96.0\% &98.9\% \\ 
\textsc{MAML~\cite{DBLP:journals/corr/FinnAL17}}  &Y&98.7$\pm$0.4\% &\textbf{99.9$\pm$0.1}\% &95.8$\pm$0.3\% &98.9$\pm$0.2\% \\ 
\textsc{\cite{DBLP:journals/corr/LiZCL17}}  &Y & \textsc{99.5$\pm$0.3\%} & \textbf{99.9$\pm$0.1\%} & 95.9$\pm$0.4\% & 99.0$\pm$0.2\%\\
\textsc{L2C~\cite{DBLP:journals/corr/abs-1711-06025}}  &N & \textsc{99.6$\pm$0.2\%} &\textsc{99.8$\pm$0.1\%} &\textsc{97.6$\pm$0.2\%} &\textsc{99.1$\pm$0.1\%} \\
\midrule 
\textsc{Ours} &N& \textbf{\textsc{99.8$\pm$0.1\%}} &\textbf{99.9$\pm$0.1\%} &\textbf{\textsc{98.2 $\pm$0.1\%}} &\textbf{\textsc{99.5$\pm$0.1\%}} \\
\hline
\bottomrule
\end{tabular}
\end{table*}

\begin{figure}
	\centering
	\includegraphics[width=0.9\textwidth]{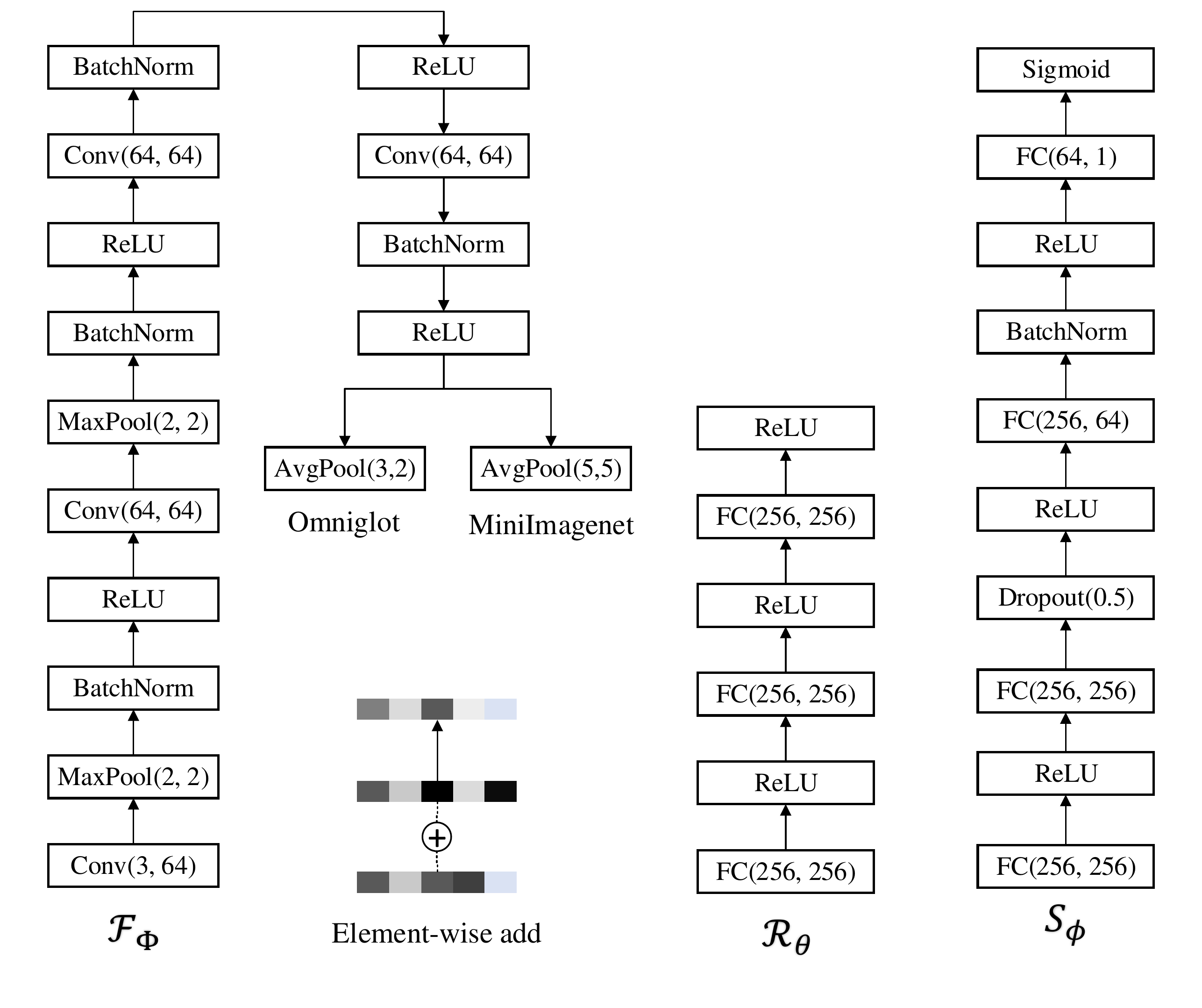}\
	\caption{Details of network architecture on Omniglot and MiniImagenet experiements.\label{fig:archo}}
\end{figure}

The detailed configuration of our networks is illustrated in Figure~\ref{fig:archo}. $Conv(k_{in}, k_{out})$ denotes a convolution layer with input channel dimension $k_{in}$ and output channel dimension $k_{out}$. All convolutional layers here have kernel size of 3x3. The numbers associated with MaxPool (resp `AvgPool') stand for the pooling \textit{kernel size} and \textit{stride size}. Previous papers~\cite{DBLP:journals/corr/abs-1711-06025,DBLP:journals/corr/SnellSZ17} resize the images to 28x28 or 20x20, which results in small feature maps from the last convolution layer, e.g. 1x1x64; In order to get a large feature map for object relation modeling, we resize the input images to 84x84. Consequently, the output from $\mathcal{F}_{\Phi}$ has 64 feature maps, each of size 7x7. Therefore, there are $(7\times 7)\times (7\times 7)=2401$ combinations, i.e object relation features, each of size $64+64=128$. The $\mathcal{R}_{\theta}$ network processes these 2401 object relation features independently through a MLP model. The output feature of each relation is of dimension 256. All features are summed over into a single feature, which is then fed into $\mathcal{S}_{\phi}$  to generate the image similarity score $s_{ab}$. All Omniglot experiments are trained with Adam~\cite{kingma2014adam} with a learning rate of 0.001 and no weight decay.

Following the experimental setting in previous papers, we compare our approach with existing methods on four tasks, namely 5-way 1-shot, 5-way 5-shot, 20-way 1-shot and 20-way 5 shot classification.   The results in terms of classification accuracy are presented in Table~\ref{tb:omniglot}. For existing methods, we copy their performance reported in the original papers or other published papers. Both meta-learning and metric-learning based approaches are compared. Meta-learning based solutions need to fine-tune the model over the support set on test dataset. For metric-learning based approaches, fine-tuning is not necessary. It may improve or decrease the performance as reported by Matching Nets~\cite{DBLP:journals/corr/VinyalsBLKW16}, shown in the 3rd and 4th rows in the table. The second column indicates whether the model is fined-tuned over the test support set or not.

Our results are averaged over 600 test episodes and are reported with 95\% confidence intervals.  The variance is also reported. We can see that our approach outperforms existing methods for 3 out of 4 tasks. Note that the accuracy of existing solutions are very high, especially for 5-way tasks. Hence, a small improvement over the state-of-the-art should be considered as significant. The improvement for 20-way tasks is clearer. 20-way tasks are more difficult than 5-way tasks as the model needs to be more discriminative to differentiate more classes. To confirm the advantage of our approach, we perform comparison against another difficult dataset in the next subsection.

In addition, We observe the training progress is rather stable. From Figure~\ref{fig:losso}, we can see that overfitting is not a problem for our algorithm although the weight decay is 0, the datasets are not large and there are many fully connected layers in $\mathcal{R}_{\theta}$ and $\mathcal{S}_{\phi}$. One possible reason is that the aggregation operation of object-level relations behaves like model ensemble, which helps prevent overfitting.

\subsection{Evaluation on MiniImagenet}

The MiniImagenet dataset~\cite{DBLP:journals/corr/VinyalsBLKW16} consists of 60,000 colour images with 100 classes sampled from ImageNet~\cite{imagenet_cvpr09}. Each class has 600 examples. We follow the partition scheme as in the original paper~\cite{DBLP:journals/corr/VinyalsBLKW16} to get 64, 16, 20 classes for training, validation and testing, respectively. We resize the images to 224x224 and do channel-wise standardization. No data augmentation is conducted. MiniImagenet is a more difficult benchmark than Omniglot because it has a larger number of classes and greater variations among the images within each class.

The configuration of $\mathcal{R}_{\theta}$ and $\mathcal{S}_{\phi}$ keep the same with Omniglot tasks as shown in Figure~\ref{fig:model}. The network of $\mathcal{F}_{\Phi}$ is almost the same as that in Figure~\ref{fig:archo} except the final average pooling layer has a larger kernel and stride size. This is to reduce the memory cost caused by large input images. $\mathcal{F}_{\Phi}$ generates 64 feature maps, each of size 10x10. Consequently, we have $(10\times 10)\times (10\times 10)=10,000$ combinations of object features. $\mathcal{R}_{\theta}$ processes the 10,000 combinations independently via a 3 layers MLP model. The output is summed over to generate a 256-d feature. The $\mathcal{S}_{\phi}$ network from Figure~\ref{fig:archo} is used again to generate the image pairs score $s_{ab}$. All MiniImagenet experiments are trained with Adam~\cite{kingma2014adam} with a learning rate of 0.001 and no weight decay.

Four tasks are conducted to do the evaluation, namely, 5-way 1-shot, 5-way 5-shot, 20-way 1-shot and 20-way 5-shot classification. In Table~\ref{tb:miniimagenet}, we report the classification accuracy including the mean and variance over 600 test episodes. The performance of existing methods are copied from their original papers or other papers. We can see that our model achieves absolute improvement over existing methods. The above observations are consistent with the results on Omniglot dataset, which indicates that our approach has larger capacity in modelling more difficult tasks.

\begin{table*}[htb]
\centering
\caption{Performance comparison on MiniImagenet dataset. \textit{FT} means fine-tuned or not.}
\label{tb:miniimagenet}
\begin{tabular}{@{} lccccc @{}}

\toprule
\hline

\textbf{MODEL} &
\textbf{FT } & 
\multicolumn{2}{c}{\textbf{5-WAY }} &
\multicolumn{2}{c}{\textbf{20-WAY }} \\

& \textbf{} & 1-shot & 5-shot & 1-shot &5-shot  \\
\hline
\midrule

\textsc{Matching Nets} & N & $43.6 \pm 0.8 \% $ & $55.3 \pm 0.7 \% $ & $17.3 \pm 0.2 \% $ & $22.7 \pm 0.2 \% $  \\ 
\textsc{Meta-LSTM} & N & $43.4 \pm 0.8 \% $ & $60.6 \pm 0.7 \% $ & $16.7 \pm 0.2 \% $ & ${26.1 \pm 0.3 \%} $\\ 
\textsc{MAML} & Y& $48.7 \pm 1.8 \% $ & $63.1 \pm 1.0 \% $ & $16.5 \pm 0.6 \% $ & $19.3 \pm 0.3 \% $ \\ 
\textsc{Meta-SGD}  & Y & $50.5 \pm 1.9 \% $ & $64.0 \pm 1.0 \% $ & $17.6 \pm 0.6 \%$ & $29.0 \pm 0.4 $\% \\
\textsc{Meta Nets} & N &49.2 $\pm$ 0.9\% & - & - & - \\ 
\textsc{Prototypical Nets} &N  &49.4 $ \pm $ 0.8\% & 68.2 $\pm$ 0.7\% & - & -  \\  
\textsc{L2C}&N& 50.44 $\pm$ 0.82\% &65.32 $\pm$ 0.7\% & - & - \\
\midrule 
\textsc{Ours} & N & \textbf{59.0 $\pm$ 1.0\%} &\textbf{70.9$\pm$ 0.5\%} &\textbf{22.2 $\pm$ 0.3\%} &\textbf{32.2$\pm$ 0.2\%} \\
\hline
\bottomrule
\end{tabular}
\end{table*}

\section{Discussion}
In this paper, we learned an object level representation for few-shot learning. 
In particular, we consider each vector $obj(\hat{x}_a, i)$ from feature map $\hat{x}_a$ of sample $x_a$ as an object. We explore the similarity of sample $x_a$ and $x_b$ on the level of $obj(\hat{x}_a, i)$ and $obj(\hat{x}_b, j)$. Then we learned a metric mapping aggregated vector $m_{ab}$ into a space such that nearest neighbor search could be applied to predict the query sample's label. 

\begin{figure*}[htb]
	\centering
	\begin{subfigure}[]{
	    \includegraphics[width=0.35\textwidth]{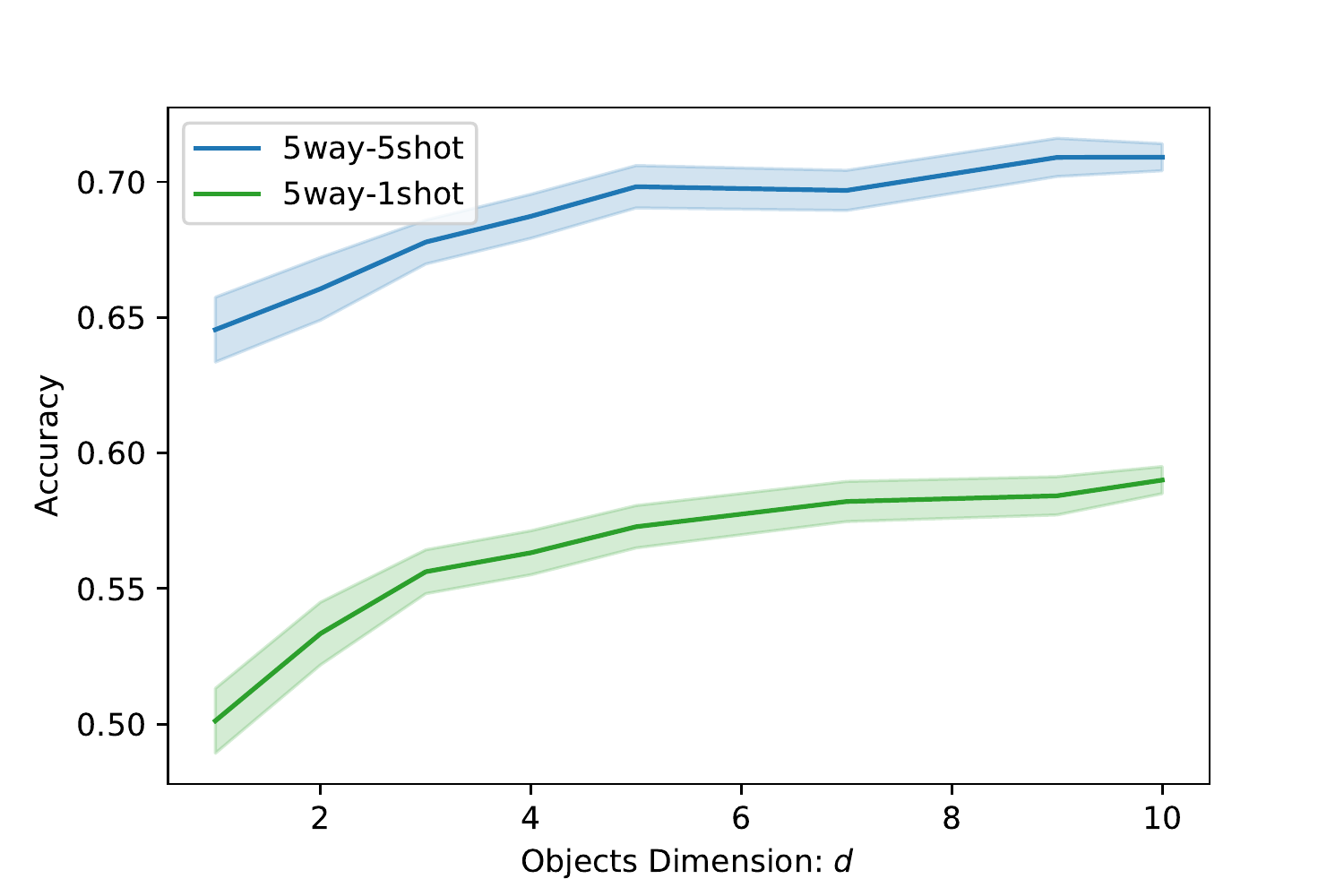}
	    \label{fig:num_obj}}
	\end{subfigure}
	\begin{subfigure}[]{
	    \includegraphics[width=0.29\textwidth]{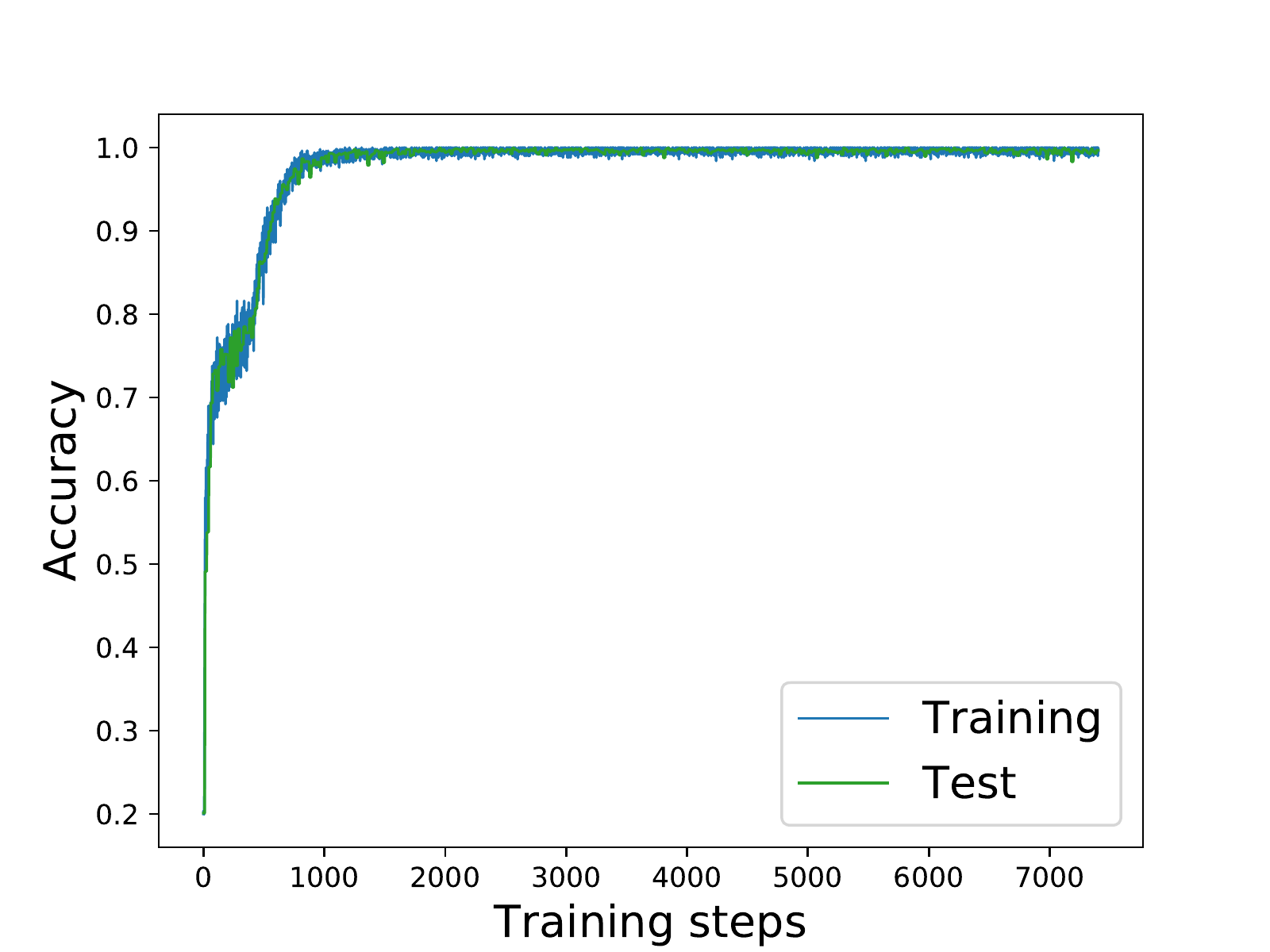}
	    \label{fig:acco}}
	\end{subfigure}
	\begin{subfigure}[]{
	\includegraphics[width=0.29\textwidth]{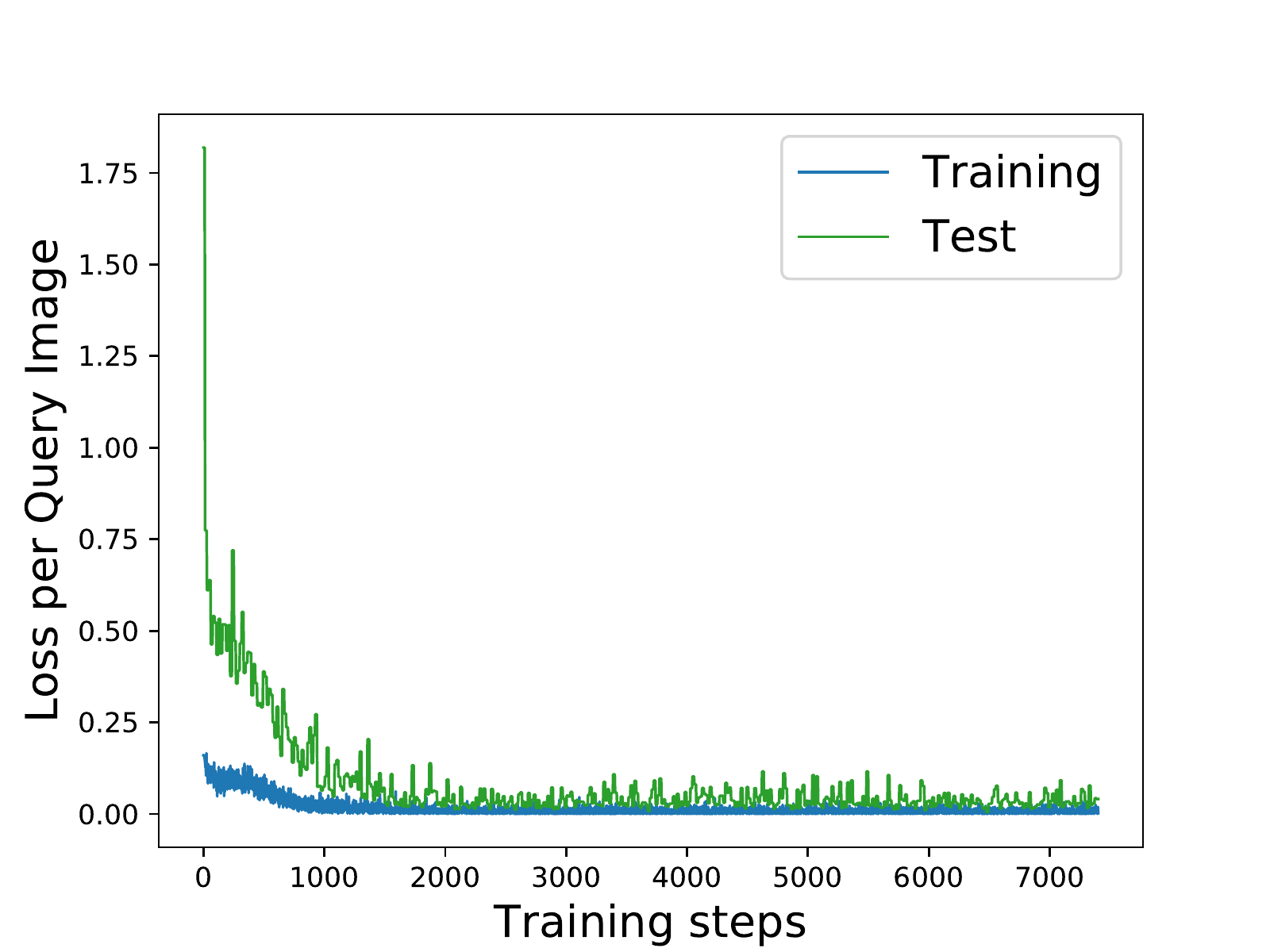}
	\label{fig:losso}}
	\end{subfigure}
	\caption{(a):Influence of objects number=$d^2$ on accuracy for MiniImagenet experiments. (b):Training and testing Accuracy curves for 5-way 5-shot Omniglot experiment. (c):Training and testing Loss curves for 5-way 5-shot Omniglot experiment.}
\end{figure*}

Our best performance on MiniImagenet experiments is achieved on dimension of objects $d=10$, which means one representation $\hat{x}_a$ includes $10*10$ number of objects. We also explored the affect on different number of objects on MiniImagenet dataset as Figure~\ref{fig:num_obj}. It demonstrates that the number of objects in representation have an important effect on performance. When $d$ is too small, e.g. $d=1$, only one objects can be learned and thus can not exploit the rich objects information. Meanwhile, When increasing the number of objects, e.g. $d=10$, the improvement become trivial. Therefore, an appropriate value of $d$ should be set properly.

In our work, we resize the image size of MiniImagenet to 224x224. Some previous work, e.g.~\cite{DBLP:journals/corr/abs-1711-06025}, resize the raw image to 84x84. We argue that our algorithm need to extract rich information from raw sample, for example, we extract $10*10$ number of objects information when $d=10$. We also observe that the image size does have very trivial influence on performance. To fair comparison, we re-run the experiments based on published source code from ~\cite{DBLP:journals/corr/abs-1711-06025}. Table~\ref{tab:rerun_l2c} indicates the performance of L2C varies subtly with 224x224 image size, even decreasing 0.28\% on 5way 1shot experiments.


\begin{table}[]
\centering
\caption{L2C experiments on MiniImagenet with different image size.}
\label{tab:rerun_l2c}
\begin{tabular}{lll}
\toprule 
\cite{DBLP:journals/corr/abs-1711-06025}     & 5way-1shot & 5way-5shot \\
\midrule
84x84   & 50.44\%    & 65.32\%    \\
224x224 & 50.16\%    & 65.98\%   \\
\bottomrule
\end{tabular}
\end{table}

\section{Related Work}


Few-shot Learning is the task of learning over datasets with few examples per category. It is useful for recognizing new categories, e.g. products. With the resurgence of deep learning, most few-shot image classifiers are based on ConvNets. A simple solution is to fine-tune the ConvNets trained on a similar dataset with many examples per category. However, the widely used gradient based optimization algorithms (e.g. mini-batch Stochastic Gradient Descent, SGD) need a lot of examples to adapt (fine-tune) the ConvNets~\cite{ravi2016optimization} for the new categories. Therefore two types of approaches are proposed recently.

\textbf{Meta Learning} Towards the optimization challenge, meta learning trains a meta learner that guides the optimization algorithms to fine-tune the learner (i.e. classifier). It is also called learning to learn. The meta learner is trained iteratively and learns slowly. For each iteration, an episode is sampled from the training dataset, which has the same setting as the test scenario. In other words, an episode has the same number of categories and the same number of examples per category as the test. The meta learner is trained to fine-tune the classifier for a large number of episodes. After training, the meta learner is expected to guide the learning of basic learner, e.g. the classifier for the test episodes. The meta learner of MAML~\cite{DBLP:journals/corr/FinnAL17} learns good parameter initialization such that the fine-tuning can adapt the parameters quickly and effectively for the test episodes. Meta-SGD's meta learner~\cite{DBLP:journals/corr/LiZCL17} generates both the initialization and learning rate for the fine-tuning optimization algorithm. A more aggressive approach~\cite{ravi2016optimization} is to learn a LSTM~\citet{Hochreiter:1997:LSM:1246443.1246450} to generate the updates for fine-tuning. It replaces the optimization of basic learner with LSTM for fine-tuning.

\textbf{Metric Learning} Another set of approaches eliminate the fine-tuning step to avoid the optimization problem. They learn a general embedding function to project both training and testing images into a metric space, where nearest neighbor search could be used as the classifier. \cite{kochsiamese} adapt Siamese network to do the feature embedding by training the network to predict the relation (from the same category or different categories) of two training images.

Meta-LSTM~\cite{DBLP:journals/corr/VinyalsBLKW16} and memory-augmented networks~\cite{DBLP:journals/corr/SantoroBBWL16} are also applied to learn the embedding space. Prototypical Network~\cite{DBLP:journals/corr/SnellSZ17} learns an embedding for each category by averaging the features of all samples in the category. L2C~\cite{DBLP:journals/corr/abs-1711-06025} is motivated by Relation network~\cite{DBLP:journals/corr/SantoroRBMPBL17}, which explores the relation between question and each objects in one image. L2C compares two images within one episode to learn their relation scores, which serve as the metric distance. However, it ignores the rich object-level information when doing the comparison between images.

\section{Conclusion}
Supervised learning has shown great success in computer vision, audio recognition with large scale dataset available. However, few-shot learning is challenging because the training algorithms of neural network, i.e. gradient based optimization algorithms, require many iterations to fine-tune the parameters over a lot of examples for new image classes. In this paper, we learn an object level representation and exploit rich object-level information to infer image similarity. We show absolute improvements on MiniImagenet dataset and state-of-the-art perofmance on Omniglot dataset. Our algorithm is intuitive, model-agnostic and keep good generalization performance. 

\bibliography{paper}

\end{document}